\documentclass[sigconf]{acmart}
\usepackage{kotex}
\usepackage{algorithm}
\usepackage{algpseudocode}
\usepackage{amsmath}
\usepackage{caption}
\usepackage{graphicx}     
\usepackage{placeins}     
\usepackage{afterpage}    
\usepackage{stfloats}     
\usepackage{booktabs} 
\usepackage{multirow} 
\usepackage{graphicx} 
\usepackage{subcaption}
\usepackage{tabularx} 
\usepackage{colortbl}
\usepackage{array}
\usepackage{placeins}
\usepackage{enumitem} 
\usepackage{caption}
\AtBeginDocument{%
  }

\copyrightyear{2025}
\acmYear{2025}
\setcopyright{cc}
\setcctype{by}
\acmConference[CIKM '25]{Proceedings of the 34th ACM International
Conference on Information and Knowledge Management}{November 10--14,
2025}{Seoul, Republic of Korea}
\acmBooktitle{Proceedings of the 34th ACM International Conference on
Information and Knowledge Management (CIKM '25), November 10--14, 2025,
Seoul, Republic of Korea}\acmDOI{10.1145/3746252.3761268}
\acmISBN{979-8-4007-2040-6/2025/11}

\begin{document}

\title{Federated Continual Recommendation}

\author{Jaehyung Lim}
\affiliation{%
  \institution{Pohang University of\\ Science and Technology}
  \city{Pohang}
  \country{Republic of Korea}}
\email{jaehyunglim@postech.ac.kr}

\author{Wonbin Kweon}
\affiliation{%
  \institution{University of Illinois  Urbana-Champaign}
  \city{Champaign}
  \state{IL}
  \country{USA}}
\email{wonbin@illinois.edu}

\author{Woojoo Kim}
\affiliation{%
  \institution{Pohang University of\\ Science and Technology}
  \city{Pohang}
  \country{Republic of Korea}}
\email{kimuj0103@postech.ac.kr}

\author{Junyoung Kim}
\affiliation{%
  \institution{Pohang University of\\ Science and Technology}
  \city{Pohang}
  \country{Republic of Korea}}
\email{junyoungkim@postech.ac.kr}

\author{Seongjin Choi}
\affiliation{%
  \institution{Pohang University of\\ Science and Technology}
  \city{Pohang}
  \country{Republic of Korea}}
\email{sjin9805@postech.ac.kr}

\author{Dongha Kim}
\affiliation{%
  \institution{Pohang University of\\ Science and Technology}
  \city{Pohang}
  \country{Republic of Korea}}
\email{dhkim0317@postech.ac.kr}

\author{Hwanjo Yu}
\authornote{Corresponding author}
\affiliation{%
  \institution{Pohang University of\\ Science and Technology}
  \city{Pohang}
  \country{Republic of Korea}}
\email{hwanjoyu@postech.ac.kr}

\renewcommand{\shortauthors}{Lim et al.}

\begin{abstract}
The increasing emphasis on privacy in recommendation systems has led to the adoption of Federated Learning (FL) as a privacy-preserving solution, enabling collaborative training without sharing user data.
While Federated Recommendation (FedRec) effectively protects privacy, existing methods struggle with non-stationary data streams, failing to maintain consistent recommendation quality over time. 
On the other hand, Continual Learning Recommendation (CLRec) methods address evolving user preferences but typically assume centralized data access, making them incompatible with FL constraints.
To bridge this gap, we introduce \textbf{Federated Continual Recommendation (FCRec)}, a novel task that integrates FedRec and CLRec, requiring models to learn from streaming data while preserving privacy.
As a solution, we propose \textbf{F$^3$CRec}, a framework designed to balance knowledge retention and adaptation under the strict constraints of FCRec. F$^3$CRec introduces two key components: \textit{Adaptive Replay Memory} on the client side, which selectively retains past preferences based on user-specific shifts, and \textit{Item-wise Temporal Mean} on the server side, which integrates new knowledge while preserving prior information.
Extensive experiments demonstrate that F$^3$CRec outperforms existing approaches in maintaining recommendation quality over time in a federated environment. Our code is available at https://github.com/Jaehyung-Lim/F3CRec-CIKM-25.
\end{abstract}




\maketitle

\section{Introduction}
\begin{figure}
    \centering
    \includegraphics[width=1\linewidth]{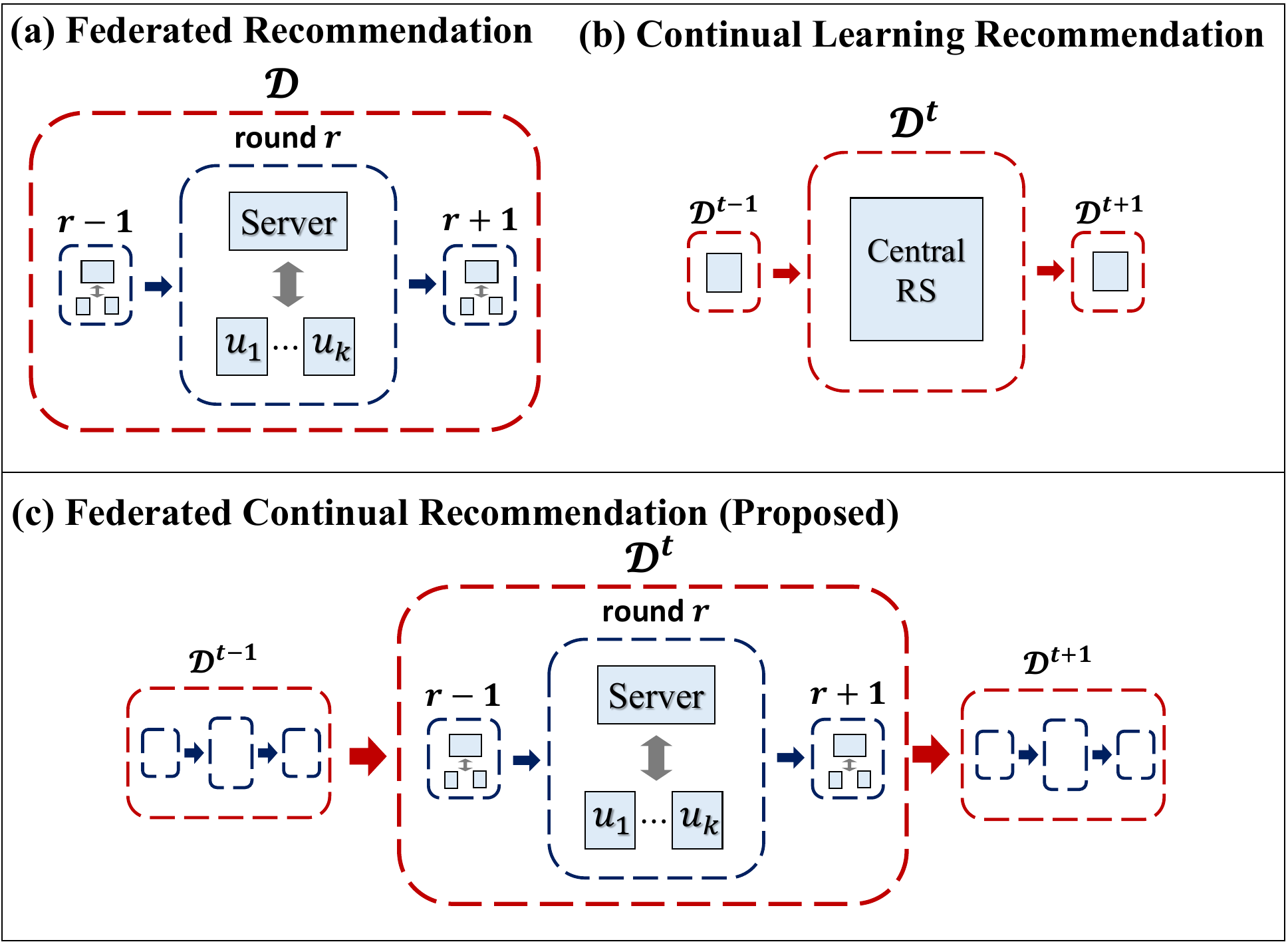}
    \caption{A conceptual comparison between (a) FedRec, (b) CLRec, and (c) FCRec (proposed).}
    \label{fig:concept_def}
\end{figure}

The extensive use of personal data in recommendation systems raises significant privacy concerns, attracting strict scrutiny from regulators and users. Regulations such as the General Data Protection Regulation (GDPR) \cite{voigt2017eu} and California Consumer Privacy Act (CCPA) \cite{pardau2018california} impose strict requirements on data collection and usage, pushing researchers and developers to find privacy-preserving approches. In response, Federated Learning (FL) has been proposed as a privacy-preserving solution, enabling collaborative model training across multiple clients without sharing their data with a central server or other clients \cite{mcmahan2017communication} (Figure \ref{fig:concept_def}a). By keeping user data local, FL reduces privacy leakage while maintaining strong performance.

When applied to recommendation systems, Federated Recommendation (FedRec) provides a promising solution for delivering personalized experiences without compromising user privacy \cite{ammad2019federated, zhang2024gpfedrec, he2024co, zhang2023dual,perifanis2022federated}.
Recent work explores enhancements in personalization, particularly in refining user and group specification, through methods such as pseudo-user relationships \cite{zhang2024gpfedrec}, adaptive item embeddings \cite{zhang2023dual, zhang2024gpfedrec}, and group-specific approaches \cite{he2024co, qu2023semi}.  
However, these approaches are inherently designed under the assumption of static, offline datasets, making them ill-suited for real-world environments where user interactions continuously evolve in a streaming fashion.
As a result, existing FedRec methods fail to address the challenge of maintaining consistent recommendation quality over time, as they overlook the balance between retaining past knowledge and adapting to new data.

On the other hand, Continual Learning Recommendation (CLRec) methods, including structure-aware regularization-based methods \cite{wang2023structure, xu2020graphsail, wang2021graph} and replay memory-based approaches \cite{cai2022reloop, zhu2023reloop2, mi2020ader}, aim to capture dynamic user preferences in the stream of user-item interactions (Figure \ref{fig:concept_def}b).
However, existing CLRec methods cannot be readily applied in FedRec environments. Structure-aware regularization methods rely on global user-item and user-user relationships \cite{xu2020graphsail, wang2021graph, wang2023structure}, and become infeasible under FedRec environments where users have access only to their local parameters and their own interaction data.
Likewise, replay memory-based approaches \cite{cai2022reloop, zhu2023reloop2, mi2020ader}, which typically rely on globally shared historical data for continual training, lose much of their effectiveness in the FedRec setting.
As a result, new methodological solutions are required to overcome these constraints and enable CLRec under the FL enviroment.

To fill the gap between FedRec and CLRec, we introduce \textbf{FCRec} (\textbf{\underline{F}}ederated \textbf{\underline{C}}ontinual \textbf{\underline{Rec}}ommendation) in Figure \ref{fig:concept_def}c, a new task designed to provide personalized recommendation on non-stationary data streams within a privacy-preserving federated setting.
Integrating both FedRec and CLRec introduces two constraints that must be satisfied simultaneously:  
(1) From FedRec, users cannot access other users' interaction data or private parameters, and the server cannot access users' private parameters or interaction data \cite{chai2020secure,zhang2023dual, perifanis2022federated, ammad2019federated}.
(2) From CLRec, training must proceed using only the currently available data, as revisiting past data is generally disallowed due to streaming scenarios and memory constraints \cite{lee2024continual, do2023continual, wang2023structure, xu2020graphsail}.  
Thus, for FCRec, a model must learn from streaming data in a federated manner without accessing past user data or other users' information.
This makes FCRec a novel and challenging task requires a new learning paradigm.  

As a solution to the proposed task, we propose \textbf{F$^3$CRec}, a \textbf{\underline{F}}rame\\work \textbf{\underline{f}}or \textbf{\underline{F}}ederated \textbf{\underline{C}}ontinual \textbf{\underline{Rec}}ommendation. 
Given the unique strategy of FedRec (i.e., client-side local training and server-side parameter aggregation), F$^3$CRec maintains a balance between new and old knowledge on both the client and server sides, within a non-stationary data stream.
To achieve this, we propose two core components: (1) \textit{Adaptive Replay Memory} on the client side to determine how much past preference information each user should retain based on individual preference shifts, 
(2) \textit{Item-wise Temporal Mean} on the server side to adaptively integrate newly learned knowledge with their previous counterparts.
By doing so, F$^3$CRec effectively adapts to evolving user preferences while preserving valuable past insights in a privacy-preserving federated environment.

Our contributions are summarized as follows:
\begin{itemize}[leftmargin=*, itemsep=2pt, topsep=3pt]
    \item We propose a new task, \textbf{FCRec} (\textbf{\underline{F}}ederated \textbf{\underline{C}}ontinual \textbf{\underline{Rec}} ommendation), which applies continual recommendation in a privacy-preserving federated recommendation setting.
    \item We introduce a novel framework, \textbf{F$^3$CRec} (\textbf{\underline{F}}ramework \textbf{\underline{f}}or \textbf{\underline{FCRec}}), which incorporates an \textit{Adaptive Replay Memory} on the client side and an \textit{Item-wise Temporal Mean} on the server side, to tackle the proposed FCRec task.
    \item We demonstrate the effectiveness of our proposed framework through extensive experiments on multiple federated backbone models and four real-world datasets.
\end{itemize}
\vspace{-10pt}

\section{Related Work}

\subsection{Federated Recommendation}
Federated Recommendation (FedRec) \cite{sun2024survey, zhang2024gpfedrec, yuan2023interaction, zhang2023dual} has emerged as a privacy-preserving solution for decentralized personalized recommendation, following the development of federated learning (FL) \cite{mcmahan2017communication}. 
Existing FedRec studies can be broadly categorized into two approaches: (1) adapting recommendation models to the federated setting, and (2) incorporating structural enhancements to address the inherent challenges in FedRec.

\vspace{+0.1cm}
\noindent \textbf{Adapting recommendation models to the federated setting.}
\cite{ammad2019federated, chai2020secure, perifanis2022federated} adapts traditional centralized recommendation models \cite{koren2009matrix, he2017neural} for federated settings. \cite{wu2022federated} incorporates GNNs under privacy constraints, though it still suffers from privacy leakage. Similarly, \cite{yan2024federated} extends FedRec with heterogeneous information networks (HINs) but still transmits user-item interactions. To mitigate cold-start and cross-domain recommendation issues, \cite{guo2024prompt, zhang2024ifedrec} integrate item attributes by leveraging content-based feature. To reduce communication costs, \cite{muhammad2020fedfast} employs active sampling, while \cite{nguyen2024towards} utilizes low-rank parameterization to minimize update transmission. 

\vspace{+0.1cm}
\noindent \textbf{Structural enhancements in federated recommendation.}
Recent studies also have explored ways to enhance FedRec by implicitly capturing user-specific preferences, or user-user relationships while maintaining privacy constraints based on aforementioned FedRec strategy \cite{chai2020secure,perifanis2022federated}.
Considering personalization, \cite{zhang2023dual} enhances user modeling with personalized item embeddings, and \cite{luo2024perfedrec++} introduces self-supervised pretraining to improve representation learning. 
\cite{he2024co, qu2023semi} utilize group-wise information to get specified parameters.
To mitigate aforementioned privacy issue from \cite{wu2022federated}, \cite{zhang2024gpfedrec} constructs relation graphs from updated item embeddings instead of users' data or private parameter.

\vspace{+0.1cm}
\noindent \textbf{Federated recommendation in data streams.}
Despite these advancements, most FedRec methods assume stationary user interactions, making them ill-suited for real-world evolving environments.

\vspace{-0.1cm}

\subsection{Continual Recommendation (CLRec)}
While traditional continual learning (CL) focuses on well-defined tasks (e.g., class \cite{dong2022federatedclassincrementallearning, zhou2024expandablesubspaceensemblepretrained} / task \cite{NEURIPS2019_3b220b43, 10543076} / domain \cite{li2024personalizedfederateddomainincrementallearning, garg2021multidomainincrementallearningsemantic}-level), recommendation systems exhibit a more fluid structure without clear task boundaries.
Instead of discrete tasks, user preferences evolve continuously, necessitating a distinct CL paradigm. 
Consequently, CLRec extends CL concepts to recommendation tasks by incrementally updating models and preserving historical information of  user-item relationships

\noindent \textbf{Structure-aware regularization-based CLRec.} 
One approach to continual learning in recommendation systems is to apply regularization methods.
In particular, structure-aware regularization methods leverage graph-based techniques to preserve topological knowledge. 
For example, \cite{xu2020graphsail} maintains knowledge at multiple levels (global, local, node), while \cite{wang2021graph} aligns node embeddings and model layers across tasks to retain past representations. More recently, \cite{wang2023structure} refines user-specific constraints to accommodate distinct preference shifts.

\noindent \textbf{Replay memory-based CLRec.} 
Additionally, replay memory-based methods are also utilized in CLRec.
\cite{mi2020ader} employs herding \cite{welling2009herding} to store pivotal samples for rehearsal, whereas \cite{cai2022reloop, zhu2023reloop2} maintain an error memory for refining future predictions. \cite{lee2024continual} incorporates stability-plasticity mechanisms to select the most informative historical samples.
Although these methods excel in centralized settings, they typically assume full access to past user data or global parameters, which raises privacy and scalability concerns.

\noindent \textbf{Continual recommendation in FL environments.} 
Structure-aware regularization methods, which leverage global user-item and user-user relationships \cite{xu2020graphsail, wang2021graph, wang2023structure}, become infeasible in FedRec environments where each user can only access their own interactions.
Similarly, replay memory-based approaches \cite{cai2022reloop, zhu2023reloop2, mi2020ader}, which typically depend on shared historical data for continual training, lose much of their effectiveness in the FedRec setting.
Thus, novel methodological solutions are required to address these constraints and facilitate CLRec in the FL environment.

\section{Proposed Task: FCRec}
Despite progress in FedRec and CLRec, their integration remains unexplored. 
This work introduces FCRec as a novel task, bridging FedRec and CLRec to ensure adaptability in real-world settings.

\subsection{Concept Definition}
\label{subsec:concept_def}
Let a sequence of temporally partitioned data blocks be denoted as  $\mathcal{D}^0, \dots, \mathcal{D}^T$ . In the context of CL, each partitioned data block $\mathcal{D}^t$, representing a set of interactions, is treated as a distinct task.
Before training on $\mathcal{D}^t$, model parameters are initialized with those from $\mathcal{D}^{t-1}$.
In the FedRec setting, each user is treated as a disjoint client, maintaining their own local models.
During each training round, the server collects trained public parameters (e.g., item embeddings) from clients and aggregates them to produce a global model update.
The aggregated parameters are then distributed back to clients.
Since these two learning paradigms are combined, the task inherits constraints from both CLRec and FedRec.

\noindent \textbf{Constraints from CLRec.}
Clients can access only their own data from the current data block, following the non-revisiting constraint in CLRec settings \cite{lee2024continual, do2023continual, wang2023structure, xu2020graphsail}, which arises from limited memory and streaming data scenarios.
To ensure fast and efficient model updates in a streaming setting, they cannot utilize data from previous time blocks \(\mathcal{D}^0, \dots, \mathcal{D}^{t-1}\) during training on $\mathcal{D}^{t}$.

\noindent \textbf{Constraints from FedRec.} 
Clients cannot access other clients' data, private parameters, or trained models at any stage.
Additionally, the global server does not have access to client-side interaction data or private parameters, aligning with FL principles \cite{zhang2024ifedrec, zhang2023dual, he2024co, perifanis2022federated, mcmahan2017communication}.  

\noindent
These constraints introduce fundamental challenges in FCRec, where models must continuously adapt to non-stationary data streams without access to centralized storage or past interactions.

\subsection{Problem Formulation}
We define $\mathcal{U}^t$ and $\mathcal{I}^t$ as the sets of users and items that have appeared in $\mathcal{D}^0, \cdots, \mathcal{D}^t$.  
For a given user $u$, the recommendation model $\mathcal{F}$ is parameterized by $\theta_u = \{\Phi_u,Q_u\}$, where $\Phi_u$ represents the user's private parameters and $Q_u$ indicates the public parameters shared with the server. 
The predicted score for item $i$ is given by $\hat{y}_{u,i} = \mathcal{F}(u, i \mid \theta_u)$.  
There are multiple choices for the parameter configuration. Following \cite{he2024co, zhang2023dual, zhang2024gpfedrec}, we define $\Phi_u$ as the user embedding, a personalized score function, or a combination of both, while $Q_u$ corresponds to the item embedding.
We focus on a common scenario where recommendations are based solely on implicit feedback from user-item interactions. Specifically, $y_{u,i} = 1$ indicates that user $u$ has interacted with item $i$, while $y_{u,i} = 0$ denotes the absence of such an interaction.

\noindent \textbf{CLRec side.}
Each dataset $\mathcal{D}^{t}$ is divided into a train/valid/test set. The model is trained on the training set and subsequently evaluated to assess how well it has learned user preferences for the corresponding block.  
Once training on $\mathcal{D}^{t}$ is completed, the learned parameters for user $u$ are denoted as $\theta_u^{t} = \{\Phi_u^{t}, Q_u^{t}\}$.  
Before starting training on $\mathcal{D}^{t+1}$, $\theta_u^{t+1}$ is initialized with $\theta_u^{t}$.

\noindent \textbf{FedRec side.} 
Each dataset $\mathcal{D}^t$ is trained over multiple rounds to gradually refine the model. 
During training, a subset of users is sampled in each training round $r$, and the selected users are denoted as $\mathcal{U}^{t,r}$.  
For a user $u \in \mathcal{U}^{t,r}$, the global server sends the public parameter $Q_g^{t,r}$, which initializes $Q_u^{t,r}$.
The user then trains on their local data $\mathcal{D}^t(u)$ using the initialized parameter $Q_u^{t,r}$ along with their private parameter $\Phi_u^{t,r}$.  
After training at round $r$, user $u (\in \mathcal{U}^{t,r})$ transmits $Q_u^{t,r}$ to the server.  
The server then aggregates the received parameters $\{Q_u^{t,r}\}_{u \in \mathcal{U}^{t,r}}$ to obtain $Q_g^{t,r+1}$.  
Then, the updated public parameter is redistributed to the users.  
\section{Proposed Method: F$^3$CRec}
\label{sec:method}
\begin{figure*}
    \centering
    \includegraphics[width=\textwidth]{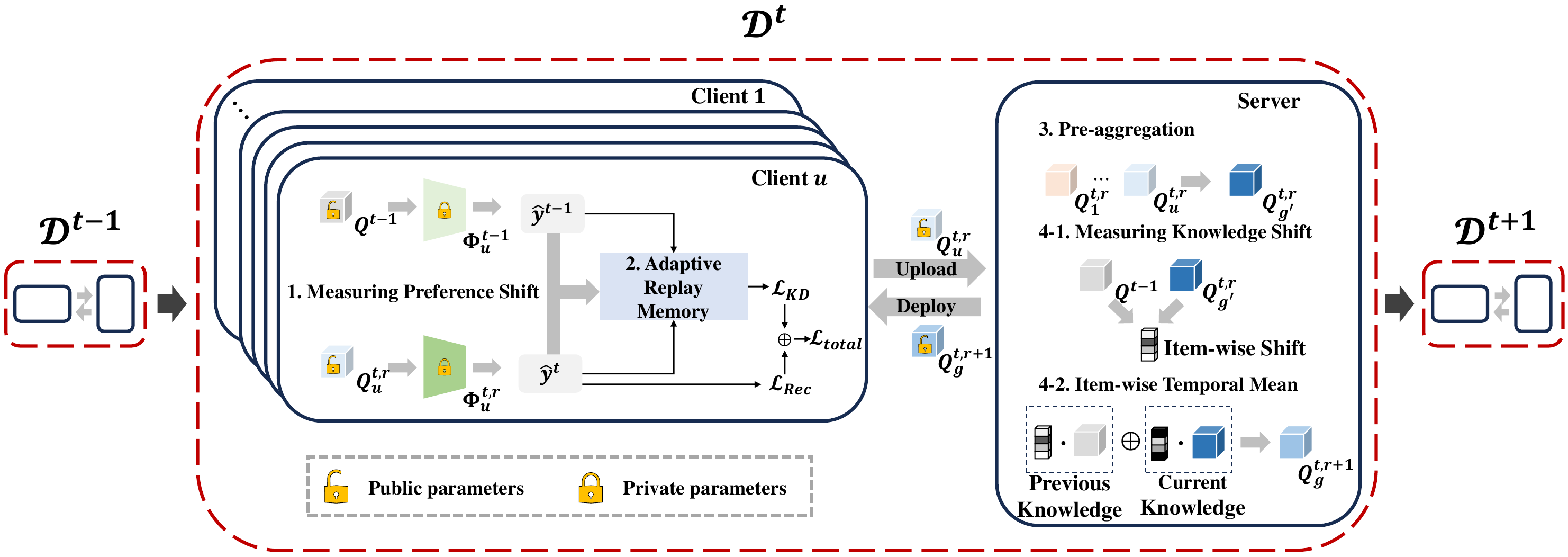}
    \vspace{+0.1cm}
    \caption{The overall framework of F$^3$CRec}
    \label{fig:method}
\end{figure*}

\subsection{Overview}
We propose \textbf{F$^3$CRec}, a novel Federated Continual Recommendation framework that applies CL at both the client and server sides in an FL setting.
F$^3$CRec is a model-agnostic framework, allowing it adaptable for various FedRec famework. To handle the non-stationary data stream under FCRec constraints, F$^3$CRec incorporates:
\begin{itemize}[leftmargin=*]
    \item \textbf{Adaptive Replay Memory} (Section \ref{subsection:client}): Adjusts knowledge retention per user by measuring ranking discrepancy between past and current preferences, ensuring adaptation without sharing interaction data or parameters across users.
    \vspace{0.1cm}
    \item \textbf{Item-wise Temporal Mean} (Section:\ref{subsection:serverside}) Aggregates item embeddings on the server while adaptively combining previous and current knowledge based on the knowledge shift of each item.
\end{itemize}
\noindent F$^3$CRec (1) adaptively transfers previous knowledge per user during local training, and (2) dynamically integrates past and current knowledge during server-side aggregation. The overall framework of our method is illustrated in Figure \ref{fig:method}.

\subsection{Client-side Continual Learning}
\label{subsection:client}
Existing CLRecs leverage global knowledge for training, but this approach conflicts with the constraints of FedRec. To address this, we propose a method where, during $\mathcal{D}^t$ for user $u$, the model updates $\theta_u^t$ using only the user's own parameters $\theta_u^{t-1}$ and the current data $\mathcal{D}^t(u)$. Notably, only $Q_u$ (or $Q_g$) is exchanged between the client and the server. At the beginning of $\mathcal{D}^t$, $\Phi_u^t$ is initialized with $\Phi_u^{t-1}$. The procedure is described in Algorithm \ref{algorithm:client}

\subsubsection{\textbf{Measuring Preference Shift}}
Effectively learning new data while preserving previously acquired knowledge requires knowledge retention and adaption \cite{wang2023structure, zhu2023reloop2}.
To achieve this, we first measure how much a user's preferences have shifted. 
This measurement, referred to as the preference shift, determines how much prior knowledge should be retained and guides the learning process for incorporating it effectively. 

Before measuring preference shift, we must define the user's preferences for a data block.
Since we cannot have access to the previous interactions, the user preference must be derived from user's each own parameters.
To this end, we assume that the top-$N$ items represent the user preference.
Accordingly, we define previous knowledge of user $u$ for $\mathcal{D}^{t-1}$ as:
\begin{equation}
\label{eq:S_u}
S_u^{t-1} = \{ i_{t_1}, \cdots, i_{t_N} \}.
\end{equation}
While training on $\mathcal{D}^t(u)$, we infer the extent of the user’s preference shift by observing changes in the rankings of items in $S_u^{t-1}$.

With $S_u^{t-1}$, we define the preference shift of user $u$ during $\mathcal{D}^t$ at training round $r$ as:
\begin{equation}
    \label{eq:Delta_u}
    \begin{aligned}
        \Delta_u^{t,r} &= \sum_{i \in S_u^{t-1}} \left| r_u^{t,r}(i) - r_u^{t-1}(i) \right| \\
        &= \sum_{k=1}^{N} \left| r_u^{t,r}(i_{t_k}) - k \right|, \quad \text{where } i_{t_k} \in S_u^{t-1},
    \end{aligned}
\end{equation}
where $r_u^{t-1}(i)$ represents the ranking of item $i$ in $S_u^{t-1}$ inferred from $\theta_u^{t-1}$, and $r_u^{t,r}(i)$ denotes its ranking of at round $r$ for $\mathcal{D}^t$.
The value $\Delta_u^{t,r}$ quantifies how much the rankings of the items in the top-$N$ list for $\mathcal{D}^{t-1}$ have changed during the training on $\mathcal{D}^t$.
This allows us to measure the overall shift in user $u$'s preferences.
A larger $\Delta_u^{t,r}$ indicates a significant change in user preferences, implying a substantial preference shift between data blocks. Conversely, a smaller $\Delta_u^{t,r}$ suggests that the user's preferences have remained consistent, making it beneficial to retain more of the previous knowledge.

\subsubsection{\textbf{Adaptive Replay Memory}}
To retain different amounts of past knowledge for each user, we propose an adaptive approach that dynamically adjusts the amount of replayed data based on the preference shift $\Delta_u^{t,r}$ .  
Unlike existing replay-memory methods \cite{mi2020ader, lee2024continual} that retain a fixed number of replay samples for all users, our approach adjusts the retention dynamically based on preference shifts.  
Users experiencing larger shifts retain fewer samples, while those with smaller shifts preserve more, ensuring that the amount of replayed knowledge is proportional to preference stability.

To achieve this, we define a user-wise consistency sampling rate based on the preference shift $\Delta_u^{t,r}$ as follows:
\begin{equation}
\label{eq:5}
    \delta_u^{t,r} = \exp(-\epsilon \cdot \Delta_u^{t,r}),
\end{equation}
where $\epsilon$ is a scaling hyperparameter. 
The value of $\delta_u^{t,r}$ is inversely proportional to $\Delta_u^{t,r}$, allowing smaller sampling rates for larger shifts and larger sampling rates for smaller shifts.
We adopt the exponential function due to its desirable properties for adaptive sampling: it is monotonic, maps non-negative inputs to the interval $(0, 1]$, and requires no additional hyperparameters or clipping operations as needed in linear or sigmoid alternatives.
Moreover, the exponential function is widely used for computing probabilities or weights that are either directly or inversely proportional to input due to its smooth decay and compatibility with Gaussian interpretation \cite{kweon2021bidirectional, lee2024continual}.

Using this sampling rate, the adaptive replay memory $M_u^{t,r}$ is constructed by sampling items from $S_u^{t-1}$ without replacement:
\begin{equation}
\label{eq:M_u}
    M_u^{t,r} = \text{SWOR}\left( S_u^{t-1}, \lfloor \delta_u^{t,r} \cdot |S_u^{t-1}| \rfloor \right),
\end{equation}
where $\text{SWOR}(A, k)$ denotes sampling $k$ items from the set $A$ without replacement.

\subsubsection{\textbf{Knowledge Distillation for Continual Learning}}
Due to the CLRec constraint (the model update should be done only with the current data block), the labels for items in $M_u^{t,r}$ are not complete.
Therefore, to incorporate the previous knowledge from $M_u^{t,r}$ into the currently trained model, we apply the knowledge distillation technique (KD) \cite{hinton2015distilling}.
We treat the model trained on $\mathcal{D}^{t-1}$ as the teacher and the model under training on $\mathcal{D}^t$ as the student.

Since we focus on the implicit feedback scenario, we adopt binary cross entropy for the KD.
The KD loss for user $u$ and item $i$ in $\mathcal{D}^t$ at round $r$ is defined as:
\begin{equation}
\label{eq:KD}
    \mathcal{L}_{KD}^{t,r}(u) = \sum_{i \in M_u^{t,r}} \left[ \hat{y}_{u,i}^{t-1} \log \hat{y}_{u,i}^{t,r} + (1 - \hat{y}_{u,i}^{t-1}) \log (1 - \hat{y}_{u,i}^{t,r}) \right],
\end{equation}
where $\hat{y}_{u,i}^{t-1}$ represents the prediction from the model trained on $\mathcal{D}^{t-1}$, while $\hat{y}_{u,i}^{t,r}$ represents the predicted value for the model being trained on $\mathcal{D}^t$ during round $r$.

\subsubsection{\textbf{Final Objective}}
Finally, the objective for user $u$ for $\mathcal{D}^t$ at round $r$ is defined as:
\begin{equation}
\label{eq:loss_total}
    \mathcal{L}^{t,r}_{total}(u) = \mathcal{L}_{Rec}^{t,r}(u) + \lambda_{KD}\mathcal{L}_{KD}^{t,r}(u),
\end{equation}
where $\lambda_{KD}$ is a hyperparameter controlling the contribution of the KD loss. 
Since we consider an implicit feedback scenario, binary cross entropy is used for $\mathcal{L}_{Rec}$.
This objective enables user $u$ to learn new preferences while appropriately retaining knowledge from previous tasks.

\subsection{Server-side Continual Learning}
\label{subsection:serverside}
In this section, we propose \textit{server-side continual learning}, a method to preserve prior knowledge at the global server level.  
Since the server has no access to user interaction data or private parameters, it rely solely on public parameters.  
During training on $\mathcal{D}^t$, the server detects knowledge shifts by comparing the public parameters aggregated from users with those obtained from $\mathcal{D}^{t-1}$.  
This enables item-wise adaptive aggregation, ensuring the retention of each item's knowledge in an adaptive manner.  
To achieve this, the server performs two key steps: \textbf{pre-aggregation} and \textbf{item-wise temporal mean}. The procedure is described in Algorithm \ref{algorithm:overall}

\subsubsection{\textbf{Pre-aggregation}}
In this step, the server aggregates public parameters trained on users' private interaction data, to compute global parameters using a naive aggregation approach:
\begin{equation}
\label{eq:preaggregation}
    Q_{g'}^{t,r} = \frac{1}{ |\mathcal{U}^{t,r}|} \sum_{u \in \mathcal{U}^{t,r}} Q_u^{t,r}.
\end{equation}
This aggregated parameter $Q_{g'}^{t,r}$ represents the pre-aggregated global parameter for round $r$ of $\mathcal{D}^{t}$.
If deployed directly to users without further processing, this would be aggregation-process of standard FedRec frameworks \cite{zhang2023dual, perifanis2022federated, zhang2024ifedrec, zhang2024fedpa}.

\begin{algorithm}[t!]
    \caption{Client-side Continual Learning}
    \label{algorithm:client}
    
    \begin{minipage}{\linewidth}
        \hspace*{0.5em} \textbf{Function:} \texttt{ClientUpdate}($u, Q_g^{t,r}, t$)
    \end{minipage}

    \begin{algorithmic}[1]
        \State Set $Q_u^{t,r} \leftarrow Q_g^{t,r}$
        \State Set $\Phi_u^{t,r}$ with latest private parameter

        \For{each epoch $e = 1,2, ..., E$}
            \For{batch $b \in \mathcal{D}^t(u)$}
                \State Compute recommendation loss $\mathcal{L}_{\text{Rec}}^{t,r}(u)$
                
                \If{$t > 0$}
                    \State Obtain $M_u^{t,r}$ using Eq.~\ref{eq:M_u}
                    \State Compute distillation loss $\mathcal{L}_{\text{KD}}^{t,r}(u)$ using Eq.~\ref{eq:KD}
                \Else
                    \State Set $\mathcal{L}_{\text{KD}}^{t,r}(u) \leftarrow 0$
                \EndIf

                \State $\mathcal{L}_{\text{total}}^{t,r}(u) \leftarrow \mathcal{L}_{\text{Rec}}^{t,r}(u) + \lambda_{\text{KD}} \cdot \mathcal{L}_{\text{KD}}^{t,r}(u)$

                \State $Q_u^{t,r} \leftarrow Q_u^{t,r} - \eta \nabla_{Q_u^{t,r}} \mathcal{L}_{\text{total}}^{t,r}(u)$ 
                \State $\Phi_u^{t,r} \leftarrow \Phi_u^{t,r} - \eta \nabla_{\Phi_u^{t,r}} \mathcal{L}_{\text{total}}^{t,r}(u)$ 
            \EndFor
        \EndFor

        \State \textbf{return} $Q_u^{t,r}$ to server
    \end{algorithmic}
\end{algorithm}
\begin{algorithm}[t!]
    \caption{F$^3$CRec – Overall Process}
    \label{algorithm:overall}
    
    \begin{minipage}{\linewidth}
        \hspace*{0.5em} \textbf{Initialize} $Q_g^0$\\
        \hspace*{0.5em} \textbf{Initialize} $\Phi_u$ for all $u \in \mathcal{U}^0$
    \end{minipage}
    
    \begin{algorithmic}[1]
        
        \For{each task $t = 0, 1, \dots, T$}
            \If{$t > 0$}
                \State Set $Q_g^t \leftarrow Q_g^{t-1}$
                \State Initialize new item embeddings: $Q_{g,i}^t$ for $i \in \mathcal{I}^{t+1} \setminus \mathcal{I}^t$
                \State Initialize new users: $\Phi_u$ for $u \in \mathcal{U}^{t+1} \setminus \mathcal{U}^t$
                
            \EndIf

            \For{each round $r = 1, 2, \dots, R$}
                \State  $\mathcal{U}^{t,r}$ $\leftarrow$ Server randomly selects a subset of clients 
                
                \For{each client $u \in \mathcal{U}^{t,r}$ \textbf{in parallel}}
                    \State $Q_u^{t,r} \leftarrow \texttt{ClientUpdate}(u, Q_g^{t,r}, t)$
                \EndFor

                \State Compute $Q_{g'}^{t,r} \leftarrow \frac{1}{|\mathcal{U}^{t,r}|} \sum_{u \in \mathcal{U}^{t,r}} Q_u^{t,r}$ 

                \If{$t = 0$}
                    \State Set $Q_g^{t,r+1} \leftarrow Q_{g'}^{t,r}$
                \Else
                    \State Obtain $Q_g^{t,r+1}$ using Eq.\ref{eq:final_temporal_mean}
                \EndIf

            \EndFor

        \EndFor
    \end{algorithmic}
\end{algorithm}

\subsubsection{\textbf{Item-wise Temporal Mean}}
Unlike client-side training where private parameters are utilized, the server does not have access to such private knowledge.
This limitation makes it impossible for the server to retain knowledge in the same way as clients, i.e., the server cannot make top-$N$ lists for users.
To overcome this, we adopt a temporal mean \cite{tarvainen2017mean}, which enables knowledge retention without relying on private parameters.
A standard temporal mean, however, applies the same weighting to all items, disregarding the fact that each item’s knowledge evolves at a different rate.

To account for this, we measure the knowledge shift for each item by comparing its embedding, which is updated based on $\mathcal{D}^{t-1}$, with the pre-aggregated embedding that is being updated based on $\mathcal{D}^t$.
New items introduced in $\mathcal{D}^t$ are excluded from this calculation. 
The knowledge shift of item $i$ is defined as:
\begin{equation}
\label{eq:phi}
    \phi_i^{t,r}  = \frac{1}{\sqrt{d}} \| Q_{g,i}^{t-1} - Q_{g',i}^{t,r} \|_2^2,
\end{equation}
where $d$ denotes the dimensionality of the item embeddings, while $Q_{g,i}^{t-1}$ and $Q_{g’,i}^{t,r}$ represent the global item embedding of item $i$ obtained from $\mathcal{D}^{t-1}$ and the pre-aggregated global embedding updated based on $\mathcal{D}^{t}$, respectively. 
We adopt the squared $L_2$ distance to quantify knowledge shift, as it is widely used in FedRec and CLRec literature (e.g., to regularize global/personalized embedding differences \cite{zhang2024gpfedrec}, or to apply embedding self-distillation \cite{xu2020graphsail}).

A larger $\phi_i^{t,r}$ indicates significant changes in the item $i$, while a smaller value suggests minimal change. Based on this, we propose the item-wise temporal mean:
\begin{equation}
\label{eq:gamma}
    \gamma_{i}^{t,r} = \frac{\beta}{1 + \phi_i^{t,r}}
\end{equation}
\begin{equation}
\label{eq:item_tempora_mean}
    Q_{g,i}^{t,r+1} = (1 - \gamma_i^{t,r}) \cdot Q_{g',i}^{t,r} + \gamma_i^{t,r} \cdot Q_{g,i}^{t-1},
\end{equation}
where $\beta \in (0,1)$ is a hyperparameter that controls the sensitivity of $\gamma_i^{t,r}$ to changes in $\phi_i^{t,r}$. Given that $\phi_i^{t,r} > 0$, the resulting coefficient $\gamma_i^{t,r}$ always lies in the range $(0,\beta]$, ensuring $\gamma_i^{t,r} < 1$. This guarantees that the update rule forms a valid convex combination for temporal mean aggregation. The adaptive weight $\gamma_i^{t,r}$ dynamically adjusts the influence of historical item embeddings based on the knowledge shift, without requiring explicit normalization.

To handle all items collectively, we define:
\begin{equation}
\label{eq:gamma_matrix}
    \gamma^{t,r} = \begin{bmatrix}
        \frac{\beta}{1 + \phi_1^{t,r}}, \cdots, \frac{\beta}{1 + \phi_{|I^{t-1}|}^{t,r}}, 0, \cdots, 0
    \end{bmatrix}^T \in \mathbb{R}^{|I^t|}.
\end{equation}
To match dimensions, zero padding is applied to $Q_g^{t-1}$:
\begin{equation}
\label{eq:zero_padding}
    Q_g^{t-1,\text{Padding}} = \text{Zero\_Padding}(Q_g^{t-1}, |I^t| - |I^{t-1}|),
\end{equation}
where $\text{Zero\_Padding}(Q, K)$ appends $K$ zero-initialized embeddings.

The final global parameter is computed as:
\begin{equation}
\label{eq:final_temporal_mean}
    Q_g^{t,r+1} = (\mathbf{1} - \mathbf{\gamma}^{t,r}) \cdot Q_{g'}^{t,r} + \mathbf{\gamma}^{t,r} \cdot Q_g^{t-1,\text{Padding}}.
\end{equation}
\section{Experiments}

We demonstrate the effectiveness of F$^3$CRec across four datasets and three backbone FedRec frameworks.
First, we present extensive experimental results showing that F$^3$CRec outperforms other CLRec methods (Section \ref{sec:exp_main_performance}). Additionally, we conduct both quantitative and qualitative analyses to validate the rationale and superiority of each proposed strategy (Sections \ref{sec:exp_client_cl} and \ref{sec:exp_server_cl}). We also perform hyperparameter sensitivity analysis under various configurations (Section~\ref{sec:hyperparameter}).
Furthermore, we perform supplementary experiments to evaluate the extent to which additional privacy-preserving considerations impact performance (Section \ref{sec:exp_privacy_preserving}). 

\begin{table}[t!]
    \scriptsize
    \centering
    \caption{Data block statistics after preprocessing}
    \label{tab:data_blocks}
    \begin{tabular}{cc|c|ccc}
        \toprule
        \multicolumn{2}{c|}{\textbf{Data Blocks}} & \textbf{$\mathcal{D}_0$} & \textbf{$\mathcal{D}_1$} & \textbf{$\mathcal{D}_2$} & \textbf{$\mathcal{D}_3$}  \\ 
        
        \midrule
        \midrule
        
        \multirow{4}{*}{\textbf{ML-100K}}
        & \# \textbf{of accumulated users}                                    & 587  & 697  & 827  & 943             \\
        & \# \textbf{of accumulated items}                                    & 1,136  & 1,146  & 1,148  & 1,152     \\
        & \# \textbf{of interactions}                                         & 58,771 & 13,060  & 13,060 & 13,062   \\ 
        & \textbf{sparsity}                                         & 91.19\% & 98.36\%  & 98.62\% & 98.80\%   \\ 
        \midrule
        
        \multirow{4}{*}{\textbf{ML-Latest-Small}}
        & \# \textbf{of accumulated users}                                    & 374     & 466  & 543  & 609          \\
        & \# \textbf{of accumulated items}                                    & 1,982   & 2,189  & 2,250  & 2,269    \\
        & \# \textbf{of interactions}                                         & 48,665  & 10,814  & 10,814  & 10,816 \\ 
        & \textbf{sparsity}                                         & 93.43\% & 98.94\%  & 99.11\% & 99.22\%   \\ 
       \midrule
        
        \multirow{4}{*}{\textbf{Lastfm-2K}}
        & \# \textbf{of accumulated users}                                    & 834  & 971  & 1,107  & 1,165        \\
        & \# \textbf{of accumulated items}                                    & 4,112 & 4,188  & 4,239  & 4,259     \\
        & \# \textbf{of interactions}                                         & 41,415 & 9,203  & 9,203  & 9,204   \\ 
        & \textbf{sparsity}                                         & 98.79\% & 99.77\%  & 99.80\% & 99.81\%   \\ 
        \midrule
        
        \multirow{4}{*}{\textbf{HetRec2011}}
        & \# \textbf{of accumulated users}                                    & 1,177  & 1,497  & 1,825 & 2,113        \\
        & \# \textbf{of accumulated items}                                    & 6,128 & 6,408  & 6,643  & 6,829     \\
        & \# \textbf{of interactions}                                         & 505,146 & 112,254  & 112,254  & 112,254   \\ 
        & \textbf{sparsity}                                         & 93.00\% & 98.83\%  & 99.07\% & 99.22\%   \\

        \bottomrule
    \end{tabular}
\end{table}

\subsection{\textbf{Experimental Setup}}
\label{sec:exp_setup}

\noindent \textbf{Datasets.} 
We employ four public real-world datasets: ML-100K \cite{harper2015movielens}, ML-Latest-Small \cite{harper2015movielens}, Lastfm-2K \cite{Cantador:RecSys2011}, and HetRec2011 \cite{Cantador:RecSys2011}, all containing timestamped interactions.
To ensure sufficient activity, we filter out users/items with fewer than 10 interactions (5 for Lastfm-2K).
To simulate non-stationary data streams, each dataset is partitioned into four blocks.
The first 60\% of interactions forms the base block ($\mathcal{D}^0$), while the remaining 40\% is chronologically divided into three incremental blocks ($\mathcal{D}^1, \mathcal{D}^2, \mathcal{D}^3$) based on chronological order following prior works \cite{lee2024continual, wang2021graph, wang2023structure, xu2020graphsail}.
For each incremental block, user interactions are randomly split into training, validation, and test sets using an 80\%/10\%/10\% ratio. Table \ref{tab:data_blocks} summarizes the detailed statistics for each block.

\noindent \textbf{Evaluation Metrics.}
During the training of each data block, we conduct evaluations on users who had interactions within that block and performed full-ranking evaluations to ensure fair comparisons. 
Similar to other CLRec frameworks, we assess performance using NDCG@20 (N@20) and Recall@20 (R@20), applying them after completing the training for the corresponding block \cite{do2023continual, lee2024continual, wang2023structure, xu2020graphsail}.

\noindent \textbf{FedRec Backbones.}
We evaluate F$^3$CRec using three representative FedRec backbones: a latent factor model \cite{chai2020secure} and two deep learning-based models \cite{perifanis2022federated, zhang2023dual}.
Specifically, we consider \textbf{FedMF} \cite{chai2020secure}, a federated extension of matrix factorization \cite{koren2009matrix} with private user embeddings and public item embeddings; \textbf{FedNCF} \cite{perifanis2022federated}, a federated adaptation of neural collaborative filtering \cite{he2017neural} where user embeddings and MLP layers are trained locally while item embeddings are collaboratively updated; and \textbf{PFedRec} \cite{zhang2023dual}, a personalized federated recommendation model that learns global item embeddings on the server and adapts them to individual data via local fine-tuning.
These models are widely used as baselines for federated recommendation \cite{zhang2023dual, zhang2024gpfedrec, he2024co}.

\begin{table*}[t!]
    \centering
    \scriptsize
    \caption{Performance comparison across four datasets and three incremental blocks. The best results are highlted in bold and second-best are underlined. \textit{Improv} (\%) represents the relative performance improvement of various methods over the result of fine tuning. OOM indicates the model ran out-of-memory. * denotes $p\leq0.05$ for the paired t-test on the best baseline on N@20.}
    \label{tab:full_performance}
     \setlength{\tabcolsep}{1.5pt}
    \resizebox{\textwidth}{!}{%
        \begin{tabular}{c|c| cccccc|cccccc|cccccc} 
        \toprule
         \multirow{2}{*}{\textbf{Dataset}} &\multirow{2}{*}{\textbf{N@20}} & \multicolumn{6}{c|}{\textbf{FedMF}} & \multicolumn{6}{c|}{\textbf{FedNCF}} & \multicolumn{6}{c}{\textbf{PFedRec}} \\
         & & FT & Reg & KD & RLP2 & SPP & \textbf{F$^3$CRec} & FT & Reg & KD & RLP2 & SPP & \textbf{F$^3$CRec} & FT & Reg & KD & RLP2 & SPP & \textbf{F$^3$CRec} \\
        \midrule
        \multirow{5}{*}{\textbf{ML-100k}} 
            & $\mathcal{D}^1$                 & 0.0794 & 0.0904 & 0.0918 & 0.0836 & 0.0878 & 0.0933      & 0.1005 & 0.1074 & 0.1039 & 0.1044 & 0.1015 & 0.1136      & 0.0957 & 0.0909 & 0.0945 & 0.0962 & 0.0942 & 0.0983 \\
            & $\mathcal{D}^2$               & 0.0768 & 0.0876 & 0.0824 & 0.0787 & 0.0838 & 0.0885      & 0.0754 & 0.0867 & 0.0808 & 0.0677 & 0.0794 & 0.0917      & 0.0881 & 0.0761 & 0.0858 & 0.0867 & 0.0902 & 0.0877 \\
            & $\mathcal{D}^3$               & 0.1002 & 0.1078 & 0.1071 & 0.1030 & 0.1117 & 0.1284      & 0.1136 & 0.1225 & 0.1222 & 0.1191 & 0.1250 & 0.1240      & 0.1137 & 0.1134 & 0.1168 & 0.1192 & 0.1231 & 0.1281 \\
            & \cellcolor{gray!15}Avg                 & \cellcolor{gray!15}0.0855 & \cellcolor{gray!15}\underline{0.0953} & \cellcolor{gray!15}0.0938 & \cellcolor{gray!15}0.0885 & \cellcolor{gray!15}0.0945 & \cellcolor{gray!15}\textbf{0.1034*}      & \cellcolor{gray!15}0.0965 & \cellcolor{gray!15}\underline{0.1056} & \cellcolor{gray!15}0.1023 & \cellcolor{gray!15}0.0971 & \cellcolor{gray!15}0.1020 & \cellcolor{gray!15}\textbf{0.1098*}      & \cellcolor{gray!15}0.0992 & \cellcolor{gray!15}0.0935 & \cellcolor{gray!15}0.0990 & \cellcolor{gray!15}0.1007 & \cellcolor{gray!15}\underline{0.1025} & \cellcolor{gray!15}\textbf{0.1047*} \\
            & \cellcolor{gray!15}\textit{Improv.}    & \cellcolor{gray!15}-- & \cellcolor{gray!15}\underline{11.49}\% & \cellcolor{gray!15}9.72\% & \cellcolor{gray!15}3.51\% & \cellcolor{gray!15}10.52\% & \cellcolor{gray!15}\textbf{21.00}\%       & \cellcolor{gray!15}-- & \cellcolor{gray!15}\underline{9.40}\% & \cellcolor{gray!15}6.04\% & \cellcolor{gray!15}0.59\% & \cellcolor{gray!15}5.69\% & \cellcolor{gray!15}\textbf{13.76}\% & \cellcolor{gray!15}-- & \cellcolor{gray!15}-5.73\%      & \cellcolor{gray!15}-0.14\% & \cellcolor{gray!15}1.54\% & \cellcolor{gray!15}\underline{3.36}\% & \cellcolor{gray!15}\textbf{5.57}\% \\

        \midrule
        \multirow{5}{*}{\textbf{ML-Latest-Small}} 
            & $\mathcal{D}^1$                 & 0.0544 & 0.0674 & 0.0656 & 0.0681 & 0.0648 & 0.0652      & 0.0567 & 0.0616 & 0.0632 & 0.0566 & 0.0634 & 0.0627      & 0.0727 & 0.0716 & 0.0785 & 0.0643 & 0.0799 & 0.0689 \\
            & $\mathcal{D}^2$               & 0.0507 & 0.0504 & 0.0659 & 0.0617 & 0.0579 & 0.0751      & 0.0466 & 0.0582 & 0.0761 & 0.0641 & 0.0716 & 0.0836      & 0.0732 & 0.0713 & 0.0687 & 0.0655 & 0.0724 & 0.0781 \\
            & $\mathcal{D}^3$               & 0.0670 & 0.0834 & 0.0609 & 0.0591 & 0.0790 & 0.0871      & 0.0731 & 0.0601 & 0.0738 & 0.0787 & 0.0796 & 0.0729      & 0.0656 & 0.0805 & 0.0744 & 0.0748 & 0.0733 & 0.0878 \\
            & \cellcolor{gray!15}Avg                 & \cellcolor{gray!15}0.0574 & \cellcolor{gray!15}0.0670 & \cellcolor{gray!15}0.0641 & \cellcolor{gray!15}0.0629& \cellcolor{gray!15}\underline{0.0672} & \cellcolor{gray!15}\textbf{0.0758*}       & \cellcolor{gray!15}0.0588 & \cellcolor{gray!15}0.0600 & \cellcolor{gray!15}0.0711 & \cellcolor{gray!15}0.0665 & \cellcolor{gray!15}\underline{0.0715} & \cellcolor{gray!15}\textbf{0.0731*}      & \cellcolor{gray!15}0.0705 & \cellcolor{gray!15}0.0745 & \cellcolor{gray!15}0.0739 & \cellcolor{gray!15}0.0682 & \cellcolor{gray!15}\underline{0.0752} & \cellcolor{gray!15}\textbf{0.0783*} \\
            & \cellcolor{gray!15}\textit{Improv.}    & \cellcolor{gray!15}-- & \cellcolor{gray!15}16.83\% & \cellcolor{gray!15}11.76\% & \cellcolor{gray!15}9.70\% & \cellcolor{gray!15}\underline{17.15}\% & \cellcolor{gray!15}\textbf{32.12}\%      & \cellcolor{gray!15}-- & \cellcolor{gray!15}2.03\% & \cellcolor{gray!15}20.89\% & \cellcolor{gray!15}13.13\% & \cellcolor{gray!15}\underline{21.69}\% & \cellcolor{gray!15}\textbf{24.33}\%                 & \cellcolor{gray!15}-- & \cellcolor{gray!15}5.59\% & \cellcolor{gray!15}4.73\% & \cellcolor{gray!15}-3.28\% & \cellcolor{gray!15}\underline{6.67}\% & \cellcolor{gray!15}\textbf{10.99\%} \\
       
        \midrule
        \multirow{5}{*}{\textbf{Lastfm-2k}} 
            & $\mathcal{D}^1$ & 0.0427 & 0.0570 & 0.0499 & 0.0464 & 0.0479 & 0.0616              & 0.0378 & 0.0476 & 0.0448 & 0.0453 & 0.0536 & 0.0469              & 0.0495 & 0.0493 & 0.0523 & 0.0522 & 0.0554 & 0.0612 \\
            & $\mathcal{D}^2$ & 0.0498 & 0.0497 & 0.0540 & 0.0545 & 0.0603 & 0.0555           & 0.0422 & 0.0490 & 0.0441 & 0.0485 & 0.0529 & 0.0536              & 0.0485 & 0.0496 & 0.0562 & 0.0554 & 0.0580 & 0.0593 \\
            & $\mathcal{D}^3$ & 0.0489 & 0.0595 & 0.0502 & 0.0597 & 0.0475 & 0.0629            & 0.0448 & 0.0399 & 0.0482 & 0.0382 & 0.0429 & 0.0514              & 0.0465 & 0.0387 & 0.0467 & 0.0483 & 0.0481 & 0.0461 \\
            & \cellcolor{gray!15}Avg & \cellcolor{gray!15}0.0471 & \cellcolor{gray!15}\underline{0.0554} & \cellcolor{gray!15}0.0513 & \cellcolor{gray!15}0.0535 & \cellcolor{gray!15}0.0519 & \cellcolor{gray!15}\textbf{0.0600*}              & \cellcolor{gray!15}0.0416 & \cellcolor{gray!15}0.0455 & \cellcolor{gray!15}0.0440 & \cellcolor{gray!15}0.0498 & \cellcolor{gray!15}\underline{0.0506} & \cellcolor{gray!15}\textbf{0.0533}               & \cellcolor{gray!15}0.0482 & \cellcolor{gray!15}0.0459 & \cellcolor{gray!15}0.0518 & \cellcolor{gray!15}0.0520 & \cellcolor{gray!15}\underline{0.0539} & \cellcolor{gray!15}\textbf{0.0555*} \\
            & \cellcolor{gray!15}\textit{Improv.} & \cellcolor{gray!15}-- & \cellcolor{gray!15}\underline{17.54}\% & \cellcolor{gray!15}8.93\% & \cellcolor{gray!15}13.58\% & \cellcolor{gray!15}10.10\% & \cellcolor{gray!15}\textbf{27.30\%} & \cellcolor{gray!15}-- & \cellcolor{gray!15}9.42\% & \cellcolor{gray!15}9.82\% & \cellcolor{gray!15}5.80\% & \cellcolor{gray!15}\underline{19.78}\% & \cellcolor{gray!15}\textbf{21.71}\%                & \cellcolor{gray!15}-- & \cellcolor{gray!15}-4.80\% & \cellcolor{gray!15}7.46\% & \cellcolor{gray!15}7.88\% & \cellcolor{gray!15}\underline{11.80}\% & \cellcolor{gray!15}\textbf{15.30}\% \\

        \midrule
        
        \multirow{5}{*}{\textbf{HetRec2011}} 
            & $\mathcal{D}^1$   & 0.0764 & 0.0825 & 0.0803 & OOM & 0.0785  & 0.0836             & 0.0696 & 0.0718 & 0.0774 & OOM & 0.0742 & 0.0817      & 0.0848 & 0.0740 & 0.0846            & OOM & 0.0880 & 0.0846 \\
            & $\mathcal{D}^2$   & 0.0716 & 0.0639 & 0.0681 & -- & 0.0668  & 0.0757            & 0.0678 & 0.0501 & 0.0708 & -- & 0.0615 & 0.0713      & 0.0791 & 0.0540 & 0.0727              & -- & 0.0663 & 0.0817 \\
            & $\mathcal{D}^3$   & 0.0941 & 0.0883 & 0.0954 & -- & 0.1004  & 0.1004              & 0.0848 & 0.0594 & 0.0945 & -- & 0.1055 & 0.0931      & 0.1047 & 0.0529 & 0.1043  & -- & 0.1020 & 0.1074 \\
            & \cellcolor{gray!15}Avg     & \cellcolor{gray!15}0.0807 & \cellcolor{gray!15}0.0782 & \cellcolor{gray!15}0.0813 & \cellcolor{gray!15}-- & \cellcolor{gray!15}\underline{0.0819}  & \cellcolor{gray!15}\textbf{0.0866*}             & \cellcolor{gray!15}0.0740 & \cellcolor{gray!15}0.0604 & \cellcolor{gray!15}\underline{0.0809} & \cellcolor{gray!15}-- & \cellcolor{gray!15}0.0804 & \cellcolor{gray!15}\textbf{0.0820*}      & \cellcolor{gray!15}\underline{0.0896} & \cellcolor{gray!15}0.0603 & \cellcolor{gray!15}0.0872  & \cellcolor{gray!15}-- & \cellcolor{gray!15}0.0855 &  \cellcolor{gray!15}\textbf{0.0913*}\\
            & \cellcolor{gray!15}Improv. & \cellcolor{gray!15}-- & \cellcolor{gray!15}-3.01\% & \cellcolor{gray!15}0.73\% & \cellcolor{gray!15}--  & \cellcolor{gray!15}\underline{1.50}\% & \cellcolor{gray!15}\textbf{7.3}\%                & \cellcolor{gray!15}-- & \cellcolor{gray!15}-18.39\% & \cellcolor{gray!15}\underline{9.26}\%               & \cellcolor{gray!15}-- & \cellcolor{gray!15}8.56\% & \cellcolor{gray!15}\textbf{10.78}\%     & \cellcolor{gray!15}-- & \cellcolor{gray!15}-32.68\% & \cellcolor{gray!15}-2.64\%               & \cellcolor{gray!15}-- & \cellcolor{gray!15}-4.58\% & \cellcolor{gray!15}\textbf{1.89}\% \\
        
        \bottomrule
        
        \toprule
         \multirow{2}{*}{\textbf{Dataset}} &\multirow{2}{*}{\textbf{R@20}} & \multicolumn{6}{c|}{\textbf{FedMF}} & \multicolumn{6}{c|}{\textbf{FedNCF}} & \multicolumn{6}{c}{\textbf{PFedRec}} \\
         & & FT & Reg & KD & RLP2 & SPP & \textbf{F$^3$CRec} & FT & Reg & KD & RLP2 & SPP & \textbf{F$^3$CRec} & FT & Reg & KD & RLP2 & SPP & \textbf{F$^3$CRec} \\
        \midrule
        \multirow{5}{*}{\textbf{ML-100k}} 

            & $\mathcal{D}^1$                 & 0.1270 & 0.1411 & 0.1512 & 0.1391 & 0.1417 & 0.1564      & 0.1934 & 0.1969 & 0.1785 & 0.1821 & 0.1851 & 0.2016      & 0.1588 & 0.1485 & 0.1539 & 0.1619 & 0.1532 & 0.1497 \\
            & $\mathcal{D}^2$               & 0.1319 & 0.1447 & 0.1422 & 0.1363 & 0.1351 & 0.1531      & 0.1380 & 0.1512 & 0.1441 & 0.1231 & 0.1415 & 0.1681      & 0.1556 & 0.1394 & 0.1418 & 0.1526 & 0.1547 & 0.1564 \\
            & $\mathcal{D}^3$               & 0.1563 & 0.1648 & 0.1625 & 0.1735 & 0.1747 & 0.1944      & 0.1686 & 0.1912 & 0.1866 & 0.1877 & 0.1922 & 0.2076                  & 0.1977 & 0.1980 & 0.1892 & 0.1941 & 0.1973 & 0.2230 \\
            & \cellcolor{gray!15}Avg                 & \cellcolor{gray!15}0.1384 & \cellcolor{gray!15}0.1502 & \cellcolor{gray!15}\underline{0.1520} & \cellcolor{gray!15}0.1496 & \cellcolor{gray!15}0.1505 & \cellcolor{gray!15}\textbf{0.1680*}      & \cellcolor{gray!15}0.1666 & \cellcolor{gray!15}\underline{0.1797} & \cellcolor{gray!15}0.1697 & \cellcolor{gray!15}0.1643 & \cellcolor{gray!15}0.1729 & \cellcolor{gray!15}\textbf{0.1924*}      & \cellcolor{gray!15}\underline{0.1707} & \cellcolor{gray!15}0.1620 & \cellcolor{gray!15}0.1616 & \cellcolor{gray!15}0.1695 & \cellcolor{gray!15}0.1684 & \cellcolor{gray!15}\textbf{0.1764*} \\
            & \cellcolor{gray!15}\textit{Improv.}    & \cellcolor{gray!15}-- & \cellcolor{gray!15}8.51\% & \cellcolor{gray!15}\underline{9.79}\% & \cellcolor{gray!15}8.12\% & \cellcolor{gray!15}8.73\% & \cellcolor{gray!15}\textbf{21.36}\%         & \cellcolor{gray!15}-- & \cellcolor{gray!15}\underline{7.87}\% & \cellcolor{gray!15}1.86\% & \cellcolor{gray!15}-1.40\% & \cellcolor{gray!15}3.78\% & \cellcolor{gray!15}\textbf{15.48}\%                  & \cellcolor{gray!15}-- & \cellcolor{gray!15}-5.11\% & \cellcolor{gray!15}-5.33\% & \cellcolor{gray!15}-0.69\% & \cellcolor{gray!15}-1.36\% & \cellcolor{gray!15}\textbf{3.32\%} \\

        \midrule
        \multirow{5}{*}{\textbf{ML-Latest-Small}} 

            & $\mathcal{D}^1$                 & 0.0724 & 0.0957 & 0.0868 & 0.0942 & 0.0933 & 0.0965      & 0.0780 & 0.0824 & 0.0906 & 0.0822 & 0.0982 & 0.1025                     & 0.1023 & 0.1170 & 0.1174 & 0.0927 & 0.1115 & 0.0964 \\
            & $\mathcal{D}^2$               & 0.0762 & 0.0771 & 0.1125 & 0.0932 & 0.0936 & 0.1126      & 0.0573 & 0.0754 & 0.1114 & 0.0824 & 0.1178 & 0.1277         & 0.1030 & 0.0924 & 0.1005 & 0.0949 & 0.1060 & 0.1224 \\
            & $\mathcal{D}^3$               & 0.1066 & 0.1269 & 0.0924 & 0.0852 & 0.1277 & 0.1158      & 0.1018 & 0.0892 & 0.1028 & 0.1162 & 0.1169 & 0.1112         & 0.0958 & 0.1116 & 0.1147 & 0.1146 & 0.1034 & 0.1149 \\
            & \cellcolor{gray!15}Avg                 & \cellcolor{gray!15}0.0851 & \cellcolor{gray!15}0.0999 & \cellcolor{gray!15}0.0972 & \cellcolor{gray!15}0.0909 & \cellcolor{gray!15}\underline{0.1049} & \cellcolor{gray!15}\textbf{0.1083*}      & \cellcolor{gray!15}0.0790 & \cellcolor{gray!15}0.0823 & \cellcolor{gray!15}0.1016 & \cellcolor{gray!15}0.0936 & \cellcolor{gray!15}\underline{0.1110} & \cellcolor{gray!15}\textbf{0.1138*}         & \cellcolor{gray!15}0.1004 & \cellcolor{gray!15}0.1070 & \cellcolor{gray!15}\underline{0.1109} & \cellcolor{gray!15}0.0974 & \cellcolor{gray!15}0.1070 & \cellcolor{gray!15}\textbf{0.1113*} \\
            & \cellcolor{gray!15}\textit{Improv.}    & \cellcolor{gray!15}-- & \cellcolor{gray!15}17.42\% & \cellcolor{gray!15}14.27\% & \cellcolor{gray!15}6.81  & \cellcolor{gray!15}\underline{23.29}\% & \cellcolor{gray!15}\textbf{27.30}\%       & \cellcolor{gray!15}-- & \cellcolor{gray!15}4.18\% & \cellcolor{gray!15}28.54\%& \cellcolor{gray!15}18.42\% & \cellcolor{gray!15}\underline{40.46}\% & \cellcolor{gray!15}\textbf{43.98}\%          & \cellcolor{gray!15}-- & \cellcolor{gray!15}6.61\% & \cellcolor{gray!15}\underline{10.48}\% & \cellcolor{gray!15}-2.97\% & \cellcolor{gray!15}6.60\% & \cellcolor{gray!15}\textbf{10.87}\% \\
        \midrule
        \multirow{5}{*}{\textbf{Lastfm-2k}} 

            & $\mathcal{D}^1$& 0.0903 & 0.0907 & 0.0909 & 0.0929 & 0.0868 & 0.0964            & 0.0814 & 0.0955 & 0.0909 & 0.0842 & 0.0925 & 0.1066             & 0.0825 & 0.0937 & 0.0830 & 0.0882 & 0.0897 & 0.1109 \\
            & $\mathcal{D}^2$& 0.0858 & 0.0802 & 0.0958 & 0.0943 & 0.1104 & 0.0997          & 0.0737 & 0.0940 & 0.0768 & 0.0868 & 0.0978 & 0.0960             & 0.0847 & 0.0867 & 0.0952 & 0.1007 & 0.0988 & 0.1048 \\
            & $\mathcal{D}^3$& 0.0825 & 0.1006 & 0.0943 & 0.0968 & 0.0918 & 0.1062          & 0.0851 & 0.0776 & 0.0938 & 0.0794 & 0.0934 & 0.0977             & 0.0931  & 0.0809 & 0.0817 & 0.1046 & 0.0925 & 0.0929 \\
            & \cellcolor{gray!15}Avg & \cellcolor{gray!15}0.0862 & \cellcolor{gray!15}0.0905 & \cellcolor{gray!15}0.0937 & \cellcolor{gray!15}0.0947 & \cellcolor{gray!15}\underline{0.0964} & \cellcolor{gray!15}\textbf{0.1008*}           & \cellcolor{gray!15}0.0801  & \cellcolor{gray!15}0.0890 & \cellcolor{gray!15}0.0872 & \cellcolor{gray!15}0.0834 & \cellcolor{gray!15}\underline{0.0945} & \cellcolor{gray!15}\textbf{0.1001*}            & 0.0868  & 0.0871 & 0.0867 & \underline{0.0978} & 0.0937 & \textbf{0.1029} \\
            & \cellcolor{gray!15}\textit{Improv.} & \cellcolor{gray!15}-- & \cellcolor{gray!15}5.02\% & \cellcolor{gray!15}8.68\% & \cellcolor{gray!15}9.86\% & \cellcolor{gray!15}\underline{11.81}\% & \cellcolor{gray!15}\textbf{16.91}\%            & \cellcolor{gray!15}-- & \cellcolor{gray!15}11.18\% & \cellcolor{gray!15}8.89\% & \cellcolor{gray!15}4.20\% & \cellcolor{gray!15}\underline{18.08}\% & \cellcolor{gray!15}\textbf{24.98}\%  & \cellcolor{gray!15}-- & \cellcolor{gray!15}0.37\% & \cellcolor{gray!15}-0.14\% & \cellcolor{gray!15}\underline{12.72}\% & \cellcolor{gray!15}7.93\% & \cellcolor{gray!15}\textbf{18.53}\% \\
        
        \midrule
        
        \multirow{5}{*}{\textbf{HetRec2011}} 

            & $\mathcal{D}^1$     & 0.1092 & 0.1068 & 0.1146 & OOM & 0.1078 & 0.1162              & 0.0930 & 0.0859 & 0.1062 & OOM & 0.1047 & 0.1070              & 0.1173 & 0.0953 & 0.1129 & OOM & 0.1241 & 0.1204  \\
            & $\mathcal{D}^2$   & 0.0999 & 0.0862 & 0.0973 & -- & 0.0966 & 0.1048              & 0.0932 & 0.0780 & 0.0950 & -- & 0.0829 & 0.0958              & 0.1116 & 0.0839 & 0.1041 & -- & 0.0902 &  0.1052 \\
            & $\mathcal{D}^3$   & 0.1086 & 0.0938 & 0.1097 & -- & 0.1114 & 0.1125              & 0.1040 & 0.0755 & 0.1046 & -- & 0.1029  & 0.1032             & 0.1086 & 0.0684 & 0.1090 & -- & 0.1086 & 0.1131 \\
            & \cellcolor{gray!15}Avg     & \cellcolor{gray!15}0.1059 & \cellcolor{gray!15}0.0956 & \cellcolor{gray!15}\underline{0.1072} & \cellcolor{gray!15}-- & \cellcolor{gray!15}0.1053 & \cellcolor{gray!15}\textbf{0.1125*}              & \cellcolor{gray!15}0.0967 & \cellcolor{gray!15}0.0798 & \cellcolor{gray!15}\underline{0.1019} & \cellcolor{gray!15}-- & \cellcolor{gray!15}0.1010 & \cellcolor{gray!15}\textbf{0.1034*}              & \cellcolor{gray!15}\underline{0.1125} & \cellcolor{gray!15}0.0825 & \cellcolor{gray!15}0.1087 & \cellcolor{gray!15}-- & \cellcolor{gray!15}0.1076 & \cellcolor{gray!15}\textbf{0.1129} \\
            & \cellcolor{gray!15}Improv. & \cellcolor{gray!15}-- & \cellcolor{gray!15}-9.72\% & \cellcolor{gray!15}\underline{1.25}\% & \cellcolor{gray!15}-- & \cellcolor{gray!15}-0.60\% & \cellcolor{gray!15}\textbf{6.26}\%                 & \cellcolor{gray!15}-- & \cellcolor{gray!15}-17.52\% & \cellcolor{gray!15}\underline{5.38}\% & \cellcolor{gray!15}-- & \cellcolor{gray!15}3.51\% & \cellcolor{gray!15}\textbf{5.45}\%                            & \cellcolor{gray!15}-- & \cellcolor{gray!15}-26.63\% & \cellcolor{gray!15}-3.38\% & \cellcolor{gray!15}-- & \cellcolor{gray!15}-4.32\% & \cellcolor{gray!15}\textbf{0.35}\% \\
        \bottomrule
        
        \end{tabular}
    }

\end{table*}

\begin{figure}[t!]
    \centering
    \includegraphics[width=0.78\columnwidth]{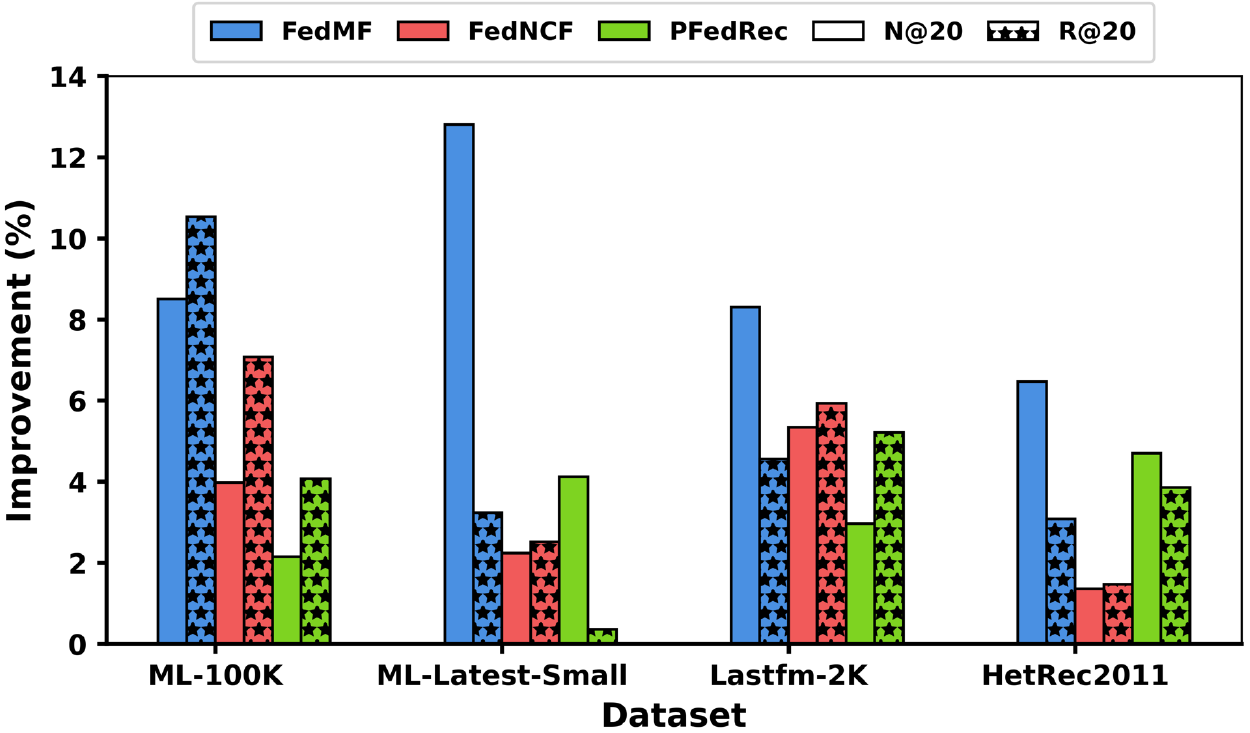}
    \Description{A comparison of F$^3$CRec and second-best performing methods across four datasets.}
    \caption{Improvement comparison of F$^3$CRec and the second-best performing methods.}
    \label{fig:best_second_improv}
\end{figure}

\noindent \textbf{Baselines.}
Since FCRec is a novel task newly proposed in this paper, there are no directly related baselines. Thus, we evaluate the performance by applying various CLRec baselines to the FedRec backbones. Specifically, we consider \textbf{FT} (Fine-Tuning) that updates the model on new data using the original loss; \textbf{Reg}, a regularization-based method that adds an MSE loss to constrain parameter deviation from the previous task; \textbf{KD}, a knowledge distillation approach that uses the Top-N item list obtained from the previous model; \textbf{RLP2} \cite{zhu2023reloop2}, a replay-based method with an error memory that stores mispredictions for self-correction, applied in FedRec with user-wise memory due to privacy constraints; and \textbf{SPP} \cite{lee2024continual} that leverages a stability-plasticity proxy and ranking discrepancy-based replay memory, with proxies applied to both private (client) and public (server) parameters in the FL setting.

\noindent \textbf{Implementation Details.}
All training and inference are conducted using PyTorch with CUDA on an RTX 3090 GPU and AMD EPYC 7313 CPU.  
We first train a base block model, then incrementally train on each data block (\(\mathcal{D}^1, \mathcal{D}^2, \mathcal{D}^3\)). Each client performs $E=1$ local epoch per round.
We use SGD optimizer, with learning rates selected per backbone: FedMF uses \{0.1, 0.5, 1\}, while FedNCF and PFedRec use [0.01,0.1] (step 0.01).
\(\lambda_{\text{KD}}\) is tuned from [1e-5,1] in powers of 10.
\(N\) is set to 30 or 50.  
\(\epsilon\) is searched as follows: ML-100K and HetRec2011 use [1e-3,9e-3]; ML-Latest-Small [1e-5,9e-5]; and Lastfm-2K [1e-4,9e-4], all with step 1e-$x$.  
\(\beta\) is tuned in [0,1] (step 0.05).  
Embedding dimension is fixed to 32 and batch size to 512. Parameter transmission is unencrypted
FedNCF and PFedRec use a 1-layer MLP.  
All baseline-specific hyperparameters follow original papers; for Reg and KD, we use same search ranges as F$^3$CRec.

\subsection{\textbf{Main Performance}}
\label{sec:exp_main_performance}
Table \ref{tab:full_performance} reports the performance of CLRec methods across multiple data blocks on four datasets, and Figure~\ref{fig:best_second_improv} compares F$^3$CRec with the second-best methods. 
Overall, F$^3$CRec shows substantial improvements across three FedRec backbones with diverse architectures and optimization strategies. 

\noindent \textbf{Misalignment of existing CLRec with FedRec.} 
Table \ref{tab:full_performance} shows that F$^3$CRec outperforms RLP2 and SPP, which adapt centralized CLRec methods to FedRec. F$^3$CRec surpasses RLP2 by 4\% to 22\%. While RLP2 leverages global knowledge for modeling preference shifts, FL constraints limit it to local knowledge, leading to performance drops. In contrast, F$^3$CRec captures preference shifts individually and item knowledge shifts globally, yielding superior performance.
SPP achieves the best baseline performance in about half of the cases, yet F$^3$CRec still outperforms SPP by 2\% to 16\%. Unlike SPP's uniform retention across users and items, F$^3$CRec uses personalized retention at the client and item-specific weighting at the server,  enhancing effectiveness.
F$^3$CRec also outperforms Reg and KD, which rely solely on user-specific data, with gains of 4\%–50\% and 1\%–33\%, respectively.
Reg applies a naive retention approach without considering preferences, whereas KD assumes equal retention for all users. F$^3$CRec explicitly accounts for individual preference shifts, leading to superior results.
Overall, existing CLRec methods are misaligned with FedRec due to neglecting decentralization. F$^3$CRec effectively addresses this by retaining knowledge on both client and server sides.

\noindent \textbf{Performance variability across FedRec models.}
While F$^3$CRec improves all backbones, FedMF and FedNCF see larger gains than PFedRec, likely due to PFedRec’s separate updates of $\Phi_u$ and $Q_u$, which may disrupt retention.  Nonetheless, F$^3$CRec consistently improves performance across all backbones.

\subsection{Closer Look on Client-side} 
\label{sec:exp_client_cl}
\begin{table}[t!]
    \scriptsize
    \caption{Results of various ablations on client-side.}
    \label{tab:ablation_client_cl}
    \resizebox{0.9\columnwidth}{!}{ 
        \begin{tabular}{ll | cc | cc}
            \toprule
            Dataset & Method & \textbf{N@20} & \textit{Decrease.} & \textbf{R@20} & \textit{Decrease.} \\ 
            \midrule
            \midrule
            \multirow{4}{*}{\textbf{ML-100k}} 
            & \textbf{F$^3$CRec}                     & \textbf{0.1034} & \textbf{--} & \textbf{0.1680} & \textbf{--} \\
            \cmidrule{2-6}
            & w/o c.c               & 0.0969 & -6.35\% & 0.1509 & -10.17\% \\
            & w/o a.r.m     & 0.0996 & -3.71\% & 0.1544 & -8.07\% \\
            & FT               & 0.0855 & -17.36\%  & 0.1384 & -17.60\% \\
            \midrule
            \multirow{4}{*}{\textbf{Lastfm-2K}} 
            & \textbf{F$^3$CRec}                     & \textbf{0.0600} & \textbf{--} & \textbf{0.1008} & \textbf{--} \\
            \cmidrule{2-6}
            & w/o c.c               & 0.0536 & -10.65\% & 0.0966 & -4.14\% \\
            & w/o a.r.m     & 0.0534 & -11.06\% & 0.0949 & -5.81\% \\
            & FT               & 0.0471 & -21.45\% & 0.0862 & -14.46\% \\
            \bottomrule
        \end{tabular}
    }
\end{table}
\begin{table}[t!]
    \small
    \centering
    \caption{Performance degradation rate on previous tasks for static and dynamic users.}
    \label{tab:analysis_client_cl}
    \resizebox{0.85\columnwidth}{!}{ 
        \begin{tabular}{l l | c c c}
        \toprule
        Dataset & Method & F$^3$CRec & w/o a.r.m & SPP \\
        \midrule
        \multirow{2}{*}{\textbf{ML-100K}} 
          & static user   & 0.5583 & 0.6674 & 0.7039 \\
          & dynamic user  & 0.7162 & 0.7552 & 0.7860 \\
        \midrule
        \multirow{2}{*}{\textbf{Lastfm-2K}} 
          & static user   & 0.2993 & 0.2772 & 0.2965 \\
          & dynamic user  & 0.3374 & 0.4567 & 0.5480 \\
        \bottomrule
        \end{tabular}
    }
\end{table}

\subsubsection{\textbf{Ablation Study on Client-side Continual Learning}}
We compare two variants: (1) \textit{w/o c.c}, which omits client-side continual learning, applying only server-side learning with item-wise temporal mean; and (2) \textit{w/o a.r.m}, which uses a fixed top-$N$ item replay memory instead of adaptively determining the replay memory size per user. In the latter, knowledge distillation applies to all top-$N$ items from the previous block.

Table \ref{tab:ablation_client_cl} shows results of the ablation study using FedMF with the ML-100K and Lastfm-2K datasets.
Excluding adaptive replay memory decreases performance, and distilling knowledge from all top-$N$ items without accounting for user preference shifts similarly reduces effectiveness.

\begin{table}[t!]
    \caption{Performance on previous task for static and dynamic users.}
    \label{tab:analysis_client_cl_value}
        \resizebox{0.85\columnwidth}{!}{ 
            \begin{tabular}{c l | c c c }
                \toprule
                Dataset & Method & F$^3$CRec &  w/o a.r.m & SPP \\ 
                \midrule
                \midrule
    
                \multirow{2}{*}{\textbf{ML-100K}}
                & static user    & \textbf{0.0788} & 0.0783 &  0.0697     \\
                & dynamic user   & \textbf{0.0620} & 0.0592 &  0.0536    \\
                
                \midrule
                \multirow{2}{*}{\textbf{Lastfm-2K}}
                & static user    & \textbf{0.0681} & \textbf{0.0681} &  0.0634     \\
                & dynamic user   & 0.0263 & 0.0334 &  \textbf{0.0368}        \\
                
                \bottomrule
            \end{tabular}
        }
\end{table}

\subsubsection{\textbf{Analysis on Adaptive Replay Memory}}
We validate whe-\\ther the proposed preference shift measure (Sec 4.2.1) effectively distinguishes static users and dynamic users.
We calculate the preference shift (Eq.\ref{eq:Delta_u}) for users with histories in both $\mathcal{D}^0$ and $\mathcal{D}^1$ after the first training round on $\mathcal{D}^1$.
Users in the bottom 20\% of preference shift are labeled static, and those in the top 20\% dynamic.
We then evaluate preference changes.
The model trained on $\mathcal{D}^0$ is first evaluated on its test set of the same block.
The model is then further trained on $\mathcal{D}^1$ and re-evaluated on the test set of $\mathcal{D}^0$ to assess knowledge preservation (Table~\ref{tab:analysis_client_cl_value}) and performance degradation (Table~\ref{tab:analysis_client_cl}), which is measured as $\frac{a_{0,0} - a_{1,0}}{a_{0,0}}$, where $a_{(t,s)}$ denotes performance when trained on $\mathcal{D}^{t}$ and evaluated on $\mathcal{D}^{s}$.
The experiments are conducted using FedMF with the N@20 metric on the ML-100K and Lastfm-2K datasets.

Results in Table \ref{tab:analysis_client_cl} and \ref{tab:analysis_client_cl_value} demonstrate that the proposed method effectively distinguishes static and dynamic users. Static users consistently achieve higher N@20 and lower preference degradation rates while dynamic users show the opposite trend. This demonstrates the effectiveness of the ranking discrepancy-based preference shift measure and highlights its capability for adaptive, user-specific continual learning leveraging solely users' own data without others.

\begin{table}[t]
    \scriptsize
    \caption{Results of various ablations on server-side.}
    \label{tab:ablation_server_cl}
    \resizebox{0.85\columnwidth}{!}{ 
        \begin{tabular}{ll | cc | cc}
            \toprule
            Dataset & Method & \textbf{N@20} & \textit{Decrease.} & \textbf{R@20} & \textit{Decrease.} \\ 
            \midrule
            \midrule
            \multirow{4}{*}{\textbf{ML-100k}} 
            & F$^3$CRec                     & \textbf{0.1034} & \textbf{--} & \textbf{0.1680} & \textbf{--} \\
            \cmidrule{2-6}
            & w/o s.c            & 0.0941 & -8.96\% & 0.1491 & -11.25\% \\
            & w/o i.t.m  & 0.0993 & -4.01\% & 0.1569 & -6.60\% \\
            & FT               & 0.0855 & -17.36\%  & 0.1384 & -17.60\% \\
            \midrule
            \multirow{4}{*}{\textbf{Lastfm-2K}} 
            & F$^3$CRec                     & \textbf{0.0600} & \textbf{--} & \textbf{0.1008} & \textbf{--} \\
            \cmidrule{2-6}
            & w/o s.c            & 0.0545 & -9.15\% & 0.0920 & -8.68\% \\
            & w/o i.t.m  & 0.0545 & -9.19 \% & 0.0961 & -4.58\% \\
            & FT               & 0.0471 & -21.45\% & 0.0862 & -14.46\% \\
            \bottomrule
        \end{tabular}
        }
\end{table}

\vspace{-0.1cm}

\subsection{Closer Look for Server-side}
\label{sec:exp_server_cl}

\subsubsection{\textbf{Ablation Study on item-wise temporal mean}}
Table \ref{tab:ablation_server_cl} shows the ablation study of server-side continual learning with FedMF on ML-100K and Lastfm-2K. We evaluate two variants: (1) \textit{w/o s.c}, where the server performs only pre-aggregation without preserving prior knowledge; and (2) \textit{w/o i.t.m}, where uniform temporal mean is applied across all items with fixed weight (i.e., $\beta$).

Both lead to performance drops, confirming the importance of item-wise temporal mean. 
In particular, \textit{w/o s.c} shows that omitting server-side retention harms performance, while \textit{w/o i.t.m} highlights the limitations of uniform temporal mean compared to item-specific weighting. These results validate the utility of item-wise temporal mean in server-side continual learning.

\begin{table}[t]
\small
    \caption{Validation of item-wise temporal mean compared to same weight aggregation with static and dynamic items.}
    \label{tab:analysis_server_cl}
        \resizebox{0.85\columnwidth}{!}{ 
        \begin{tabular}{ll | c c c}
            \toprule
            Dataset & Type & static items & dynamic items & \textit{diff} \\ 
            \midrule
            \midrule
            
            \multirow{2}{*}{\textbf{ML-100k}} 
            & F$^3$CRec                        & 0.0312 & 0.3038 & \textbf{0.2726} \\
            & w/o i.t.m     & 0.0470 & 0.2954 & 0.2484 \\
            \midrule
            
            \multirow{2}{*}{\textbf{Lastfm-2K}} 
            & F$^3$CRec                      & 0.0516 & 0.5403 & \textbf{0.4887} \\
            & w/o i.t.m     & 0.0691 & 0.3305 & 0.2614 \\
            \bottomrule
        \end{tabular}
        }
\end{table}

\subsubsection{\textbf{Analysis on Item-wise Temporal Mean}}
Table \ref{tab:analysis_server_cl} validates the effectiveness of item-wise temporal mean (Sec 4.3.2) with FedMF. 
Static and dynamic items are defined as the bottom and top 20\% in $\phi_i$ (Eq.~\ref{eq:phi}), respectively. 
We evaluate users involved in training on $\mathcal{D}^{t-1}$ but not on $\mathcal{D}^{t}$, enabling fair evaluation.
Dynamic items with high $\phi_i$ are expected to show larger shifts in collaborative filtering (CF) signals compared to the previous data block, while static items remain stable. 
To verify this, we measure the rate of change in item rankings, calculated as $|rank_{u,i}^{t-1} - rank_{u,i}^{t}| / rank_{u,i}^{t-1}$, for users who are trained on $\mathcal{D}^{t-1}$ but not on $\mathcal{D}^{t}$. The reported metric is averaged over three consecutive transitions: $\mathcal{D}^{1}$ to $\mathcal{D}^{0}$, $\mathcal{D}^{2}$ to $\mathcal{D}^{1}$, and $\mathcal{D}^{3}$ to $\mathcal{D}^{2}$.

Table \ref{tab:analysis_server_cl} shows that static items consistently have lower ranking change rates than dynamic ones, supporting the use of $\frac{1}{1 + \phi_i}$ for distinguishing them. 
Compared to uniform temporal mean, item-wise temporal mean yields lower change rates for static items and higher for dynamic ones, indicating better preservation of stable CF signals and responsiveness to dynamic shifts. Lastly, the larger gap in ranking change rates between static and dynamic items under adaptive aggregation highlights its effectiveness in adjusting CF signal reflection and preserving user preferences.

\subsection{Hyperparameter Study}
\label{sec:hyperparameter}
    
    
     
    

\begin{figure}
    \centering
    \includegraphics[width=1.0\columnwidth]{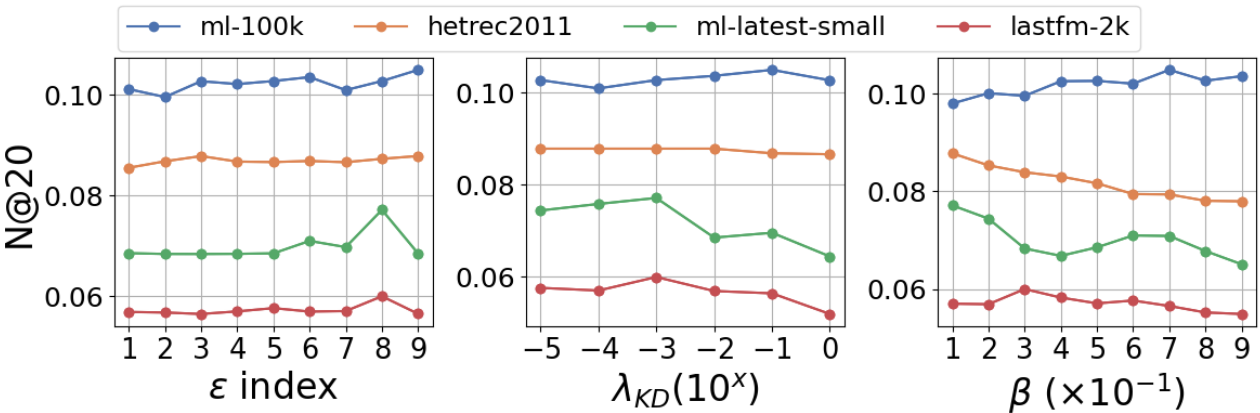}
    \caption{Effects of FedMF's hyperparameters, $\epsilon$, $\lambda_{KD}$, and $\beta$ on N@20 across four datasets.}
    \label{fig:hyperparameter_study}
\end{figure}
Figure \ref{fig:hyperparameter_study} illustrates the effects of hyperparameters ($\epsilon$, $\lambda_{KD}$, $\beta$) on FedMF performance (N@20), averaged over $\mathcal{D}^{1}$-$\mathcal{D}^{3}$.
The sampling factor $\epsilon$, which controls the amount of retained past data, is dataset-specific in scale (ML-100K, HetRec2011: $10^{-3}$; Lastfm-2K: $10^{-4}$; ML-Latest-Small: $10^{-5}$). While ML-100K exhibits irregular trends, other datasets show improved performance with larger $\epsilon$, indicating that reducing reliance on past data aids adapting preference shifts.
The KD coefficient $\lambda_{KD}$ improves performance at moderate values but degrades beyond certain thresholds—except for ML-100K—suggesting that overemphasis on prior knowledge hinders adaptation to new data. 
The parameter $\beta$, regulating retention at the server, benefits ML-100K with stable preferences, but higher values harm performance in datasets with greater shifts (ML-Latest-Small, Lastfm-2K, and HetRec2011), where reduced retention enhances adaptability.
Overall, hyperparameter tuning has a modest impact, typically varying performance by less than 0.01 in N@20. Thus, maintaining hyperparameters within reasonable ranges ensures stable performance without extensive optimization, balancing stability and adaptability in federated continual recommendation scenarios.

\begin{table}[t]
    \small
    \caption{Performance with various noise intensity $\lambda$}
    \label{tab:analysis_ldp}
    \resizebox{\columnwidth}{!}{ 
    \begin{tabular}{ll | cccccc}
        \toprule
         \textbf{Dataset} & \textbf{Noise ratio} & $\lambda=0$  & $\lambda=0.1$ & $\lambda=0.2$ & \textbf{$\lambda=0.3$} & $\lambda=0.4$ & $\lambda=0.5$ \\
         
         \midrule
         \midrule
         
         \multirow{2}{*}{\textbf{ML-100K}}
          & NDCG@20  & \textbf{0.1034} & 0.1021 & 0.0959 & 0.0951 & 0.1004 & 0.0950 \\
          & Recall@20  & \textbf{0.1680} & 0.1661 & 0.1584 & 0.1620 & 0.1633 & 0.1643\\
                
         \midrule
         
         \multirow{2}{*}{\textbf{ML-Latest-Small}}
          & NDCG@20  & \textbf{0.0758} & 0.0613 & 0.0632 & 0.0658 & 0.0602 & 0.0614 \\
          & Recall@20  & \textbf{0.1083} & 0.0870 & 0.0945 & 0.0936 & 0.0776 & 0.0836 \\

         \midrule
         
         \multirow{2}{*}{\textbf{Lastfm-2K}}
          & NDCG@20  & \textbf{0.0600} & 0.0530 & 0.0526 & 0.0516 & 0.0501 & 0.0493 \\
          & Recall@20  & 0.1008 & 0.0978 & 0.0969 & \textbf{0.1018} & 0.0929 & 0.0922 \\

          \midrule
         
          \multirow{2}{*}{\textbf{HetRec2011}}
          & NDCG@20  & \textbf{0.0872} & 0.0829 & 0.0854 & 0.0811 & 0.0824 & 0.0794\\
          & Recall@20  & \textbf{0.1105} & 0.1055 & 0.1055 & 0.1019 & 0.1024 & 0.1053\\
        
        \bottomrule
    \end{tabular}
    }
    
\end{table}

\subsection{Privacy-Preserving}
\label{sec:exp_privacy_preserving}
Although private parameters and interaction data remain private in FL settings, information leakage may occur when transmitting public parameters to the server. 
Given a user $u$ with $n_u^+$ positive and $n_u^-$ negative items, the server can compare $Q_{g}^{t,r-1}$ and $Q_u^{t,r}$ to estimate the probability of an item being positive as $n_u^+ / (n_u^+ + n_u^-)$, compromising privacy.
To mitigate this, we add zero-mean Laplace noise to user updates before transmission:
$Q_u^{t,r} = Q_u^{t,r} + \text{Laplace}(0, \lambda)$,
where $\lambda$ controls the noise intensity. While not a formal local differential privacy (LDP) mechanism, this strategy is loosely motivated by LDP principles \cite{zhang2024gpfedrec, zhang2023dual, 8416855}.
We evaluate the performance of F$^3$CRec with FedMF under varying noise intensities $\lambda \in \{0, 0.1, 0.2, 0.3, 0.4, 0.5\}$ across four datasets, averaged over all data blocks (Table~\ref{tab:analysis_ldp}).
Results show that even with stronger noise, performance degradation remains low, indicating our method reduces privacy leakage during aggregation without sacrificing utility.

\section{Conclusion}
We argue that existing CLRec methods are misaligned with the FL setting and that no prior work explicitly addresses this gap.  
To this end, we propose a new task, FCRec, and present F$^3$CRec—the first framework designed for FCRec.
F$^3$CRec incorporates two key components: Adaptive Replay Memory, which leverages user-specific preference shifts for selective client-side retention, and Item-wise Temporal Mean, which adaptively aggregates item embeddings at the server for task-aware knowledge preservation.
Extensive experiments confirm the superiority of F$^3$CRec under federated privacy constraints, and further analysis highlights its effectiveness in capturing dynamic, user- and item-level shifts.
\section{Acknowledgement}
This work was supported by IITP grant funded by the MSIT (No.2018-0-00584, RS-2019-II191906), the NRF grand funded by the MSIT (South Korea, No. RS-2024-00335873, RS-2023-00217286).

\section{GenAI Usage Disclosure}
To enhance the manuscript’s clarity and style, we used a generative language model (GPT 4o) to assist with grammar correction, academic expression, and sentence fluency, based on the authors' original draft.
However, all core ideas, including the main (technical) contributions, method design, experimental setup, and analysis, were entirely conceived and developed by the authors.
All figures, including method diagrams and experimental results, were created solely by the authors without any assistance from generative models.

\bibliographystyle{ACM-Reference-Format}
\bibliography{contents/references}

\end{document}